\pdfoutput=1

\documentclass[11pt]{article}

\usepackage{acl}

\usepackage{times}
\usepackage{latexsym}

\usepackage[T1]{fontenc}

\usepackage[utf8]{inputenc}

\usepackage{microtype}

\usepackage{inconsolata}

\usepackage{multirow} 
\usepackage{array}
\usepackage{booktabs}
\usepackage{graphicx}
\usepackage{amsmath}
\usepackage{CJKutf8}
\usepackage{url}
\usepackage{tikz-qtree}

\title{Cross-domain Chinese Sentence Pattern Parsing}

\author{Jingsi Yu$^{1}$ \ 
  Cunliang Kong$^{2}$\ 
  Liner Yang\thanks{Corresponding author.}$^{1}$ \ 
  Meishan Zhang$^{3}$ \\
 \textbf{Lin Zhu}$^{1}$  \ 
 \textbf{Yujie Wang}$^{4}$  \ 
 \textbf{Haozhe Lin}$^{2}$  \ 
 \textbf{Maosong Sun}$^{2}$ \ 
 \textbf{Erhong Yang}$^{1}$ \\
  $^1$Beijing Language and Culture University, China \quad $^2$Tsinghua University, China \\ 
  $^3$Harbin Institute of Technology (Shenzhen), China \quad 
  $^4$Beijing Jiaotong University, China \\
}

\begin{document}
\begin{CJK*}{UTF8}{gbsn}
\maketitle
\begin{abstract}
Sentence Pattern Structure (SPS) parsing is a syntactic analysis method primarily employed in language teaching.
Existing SPS parsers rely heavily on textbook corpora for training, lacking cross-domain capability.
To overcome this constraint, this paper proposes an innovative approach leveraging large language models (LLMs) within a self-training framework.
Partial syntactic rules from a source domain are combined with target domain sentences to dynamically generate training data, enhancing the adaptability of the parser to diverse domains.
Experiments conducted on textbook and news domains demonstrate the effectiveness of the proposed method, outperforming rule-based baselines by 1.68 points on F1 metrics.
\end{abstract}

\section{Introduction}
\begin{figure}[t]
    \centering
    \includegraphics[width=0.9\linewidth]{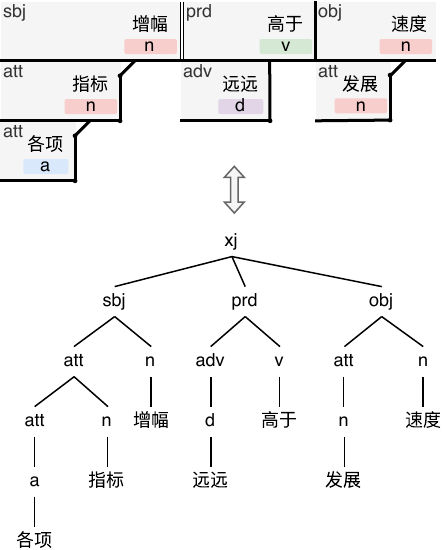}
    \caption{The SPS diagram and corresponding tree for the sentence ``\textit{各项指标增幅远远高于发展速度\ (The growth rates of all indicators far exceed the development speed)}''. These two illustrations are essentially equivalent.}
    \label{sps-exp}
\end{figure}

Syntax analysis plays a vital role in the field of natural language processing, with the primary objective of enabling models to understand syntactic information \citep{finkel:08, yu:15, bai:23} or apply it to other downstream tasks \citep{zhangyuhao:18, yan:19,liyafu:23}. 
Typical syntax analysis tasks involve constituency parsing, which emphasizes combinatory relationships between words \citep{xue:05}, and dependency parsing, which highlights modifying relationships between words \citep{lucien:59}.
In contrast to these tasks, this paper focuses on sentence pattern structure (SPS) parsing, which is mainly applied in language teaching.
As in figure \ref{sps-exp}, SPS annotates the main components (subject, predicate, object, etc.) and modifying elements (attributive, adverbial, etc.), helping language learners understand sentences \citep{pengbook:21}.

Current research on SPS focuses predominantly on improving the performance of parsers \citep{peng:14}.
Its limitation lies in the heavy reliance on textbook corpora for training, which hinders its ability to generalize well on other domains.
Efforts have been made to extend SPS treebank resources to the news domain \citep{zhangyinbing:18, xie:23}, which involves transforming constituency trees from the news domain into SPS trees through rule-based conversions.
This approach requires a substantial volume of manually or automatically annotated constituency trees in the target domain to facilitate data expansion.
However, in domains lacking such annotated data, this approach fails to effectively address the issue of cross-domain SPS parsing.

Self-training-based unsupervised domain adaptation has emerged as a promising approach for cross-domain transfer \citep{yu:15, sachan:18, he:19, rotman:19, ramponi:20,ye:20, wang:21}.
This method utilizes a source domain parser to automatically label a large-scale raw corpus from the target domain during each iteration.
High-confidence pseudo-data are then selected as additional training data to improve target domain performance.
However, the quality and quantity of raw corpus cannot always be guaranteed for low-resource domains\citep{steedman:03, qiu:14,peng:21}, which limits the use of self-training approaches.

To address the issue of cross-domain SPS parsing and broaden its application in language teaching, this paper explores the utilization of the generative abilities of large language models (LLMs).
We dynamically embed LLMs into the iterative process of self-training, enhancing their adaptability and flexibility.
Specifically, we extract partial syntactic rules from annotated data in the source domain and combine them with a small subset of sentences from the target domain.
These combined inputs are fed into an LLM to generate sentences in the target domain.
Because these sentences adhere to specific syntactic rules, they can be used for specialized training in the target domain.
Furthermore, to address the instability and hallucination issues of LLMs, we incorporated rule-based methods in both data generation and pseudotree selection.

To validate our proposed method, we conducted experiments using textbooks as the source domain and news as the target domain.
The experimental results indicate that our proposed LLM-enhanced self-training method outperforms the rule-based baselines, with an improvement of 1.68 points in the F1 metrics.

The contributions of this paper can be summarized as follows:
\begin{itemize}
    \item We present the novel task of cross-domain SPS parsing, which can effectively expand the applicability of SPS.
    \item We propose an innovative LLM-enhanced self-training method, facilitating the transfer of an SPS parser from the source domain to the target domain.
    \item Experimental findings validate the effectiveness of the proposed method, showing significant improvements in accuracy across different domains. The associated data and code will be made publicly available on \url{https://github.com}. 
\end{itemize}

\section{Related Work}

\subsection{Cross-domain Syntactic Parsing}

Syntactic parsing \citep{collins:97, mcclosky:06, finkel:08, zhu:13, dyer:15, dozat:16, stern:17, gaddy:18, guo:23} is the automatic analysis of the syntactic structure of natural language. It is a long-standing traditional NLP task, has evolved over the years. 
Research has explored the application of parser combinations \citep{mcclosky:10} and LLM-enhanced \citep{li:23} for cross-domain constituent parsing.
Furthermore, cross-domain parsing has also been investigated on other grammar formalisms, specifically dependency syntax \citep{blodgett:18, li:19, rotman:19}.
For SPS parsing, the work by \citet{xie:23} engaged with the issue of cross-domain applicability but fell short of providing a comprehensive solution. Their methodology, which is based on rule-based mapping, specifically designs mapping rules from constituent trees to SPS trees for Penn Chinese Treebank (CTB). However, this approach is not adaptable for application across various domains, limiting its generalizability.

\subsection{Automatic Conversion of Treebank}

Research on SPS grammar is relatively limited in the NLP field, while wider application of constituent treebanks and dependency treebanks. The Language and Text Resources Research Center at Beijing Normal University has constructed an STB from textbook, while other areas experience a lack of such corpora. Due to the high cost and time-consuming of manual annotation, \citet{zhangyinbing:18} and \citet{xie:23} adopted a method of treebank conversion. This involves utilizing existing treebank resources and identifying the mapping relationships between two forms of grammars to convert an existing resource treebank into target treebank. Theoretically, although different types of treebanks differ in their syntactic presentation, they essentially describe the syntactic structure of gold texts, making the conversion between different treebanks feasible. Although automatic treebank conversion has not achieved true cross-domain capabilities, the syntactic rules in the data generated by rule-based method 156
have significant utility in filtering raw corpus produced by LLM. 

\subsection{Self-training}

Self training \citep{yarowsky:95, mcclosky:06, yu:15, ramponi:20, guo:23} is a semi-supervised learning technique mainly utilized in scenarios with a limited amount of labeled data complemented by a substantial volume of unlabeled data. 
This method initiates with the training of a foundational model using the available labeled data, and then iteratively generates pseudo-data with high confidence, incrementally enlarging the dataset for model training. 
This process aims to enhance the model's performance and generalization capability in the context of scarce labeled data. 
Self-training has been extensively applied across various NLP tasks \citep{dong:11, ye:20, wang:21, yu:15, rotman:19, guo:23, he:19, sachan:18, li:23}. 
In this work, we focus on unsupervised cross-domain SPS parsing, investigating the LLM-enhanced self-training guided by rules.

\subsection{LLMs Parsing}

Due to the remarkable generative capabilities of LLMs, they have achieved success in numerous NLP tasks \citep{brown:20, he:23, mysore:23, zhang:23, pangakis:23, ashok:23, bai:23}, including commonsense reasoning \citep{zhang:22}, text summarization \citep{goyal:22, chen:23}, and massive multitask language understanding \citep{hendrycks:21}.
Research has combined LLMs with constituency parsing, demonstrating the efficacy of LLMs in syntactic parsing. \citet{li:23} employed LLMs for data augmentation, generating raw corpora across five domains and dynamically incorporating them into the training process. \citet{bai:23} assessed the performance of LLMs under zero-shot, few-shot, and full-training settings, evaluating both in-domain and out-of-domain performance.

\section{Method}

\subsection{Rule-based Mapping}

Research on constituency parsing is relatively mature, with parsers achieving high accuracy. Consequently, \citet{xie:23} designed rules for the CTB to convert constituent trees into SPS trees. We first generate constituent trees for our testing set with CoreNLP, then apply rule-based mapping to obtain SPS trees for comparative analysis.

\subsection{Berkeley Neural Parser}

The Berkeley Neural Parser \citep{kitaev:18} is a state-of-the-art tool in syntactic analysis, leveraging neural networks to accurately generate parse trees from sentences. It is distinguished by its ability to adapt across languages and domains, thanks to its integration with pre-trained language models. Its architecture, which combines a self-attentive encoder with a chart-based decoder, allows for efficient and reliable parsing. This has made the parser a preferred choice for syntactic parsing tasks among researchers and practitioners alike.
Based on the bert-base-chinese model, we continue training with textbook data and news data after rule-based mapping, serving as our foundational baseline.

\begin{figure}[t]
    \centering
    \includegraphics[width=1\linewidth]{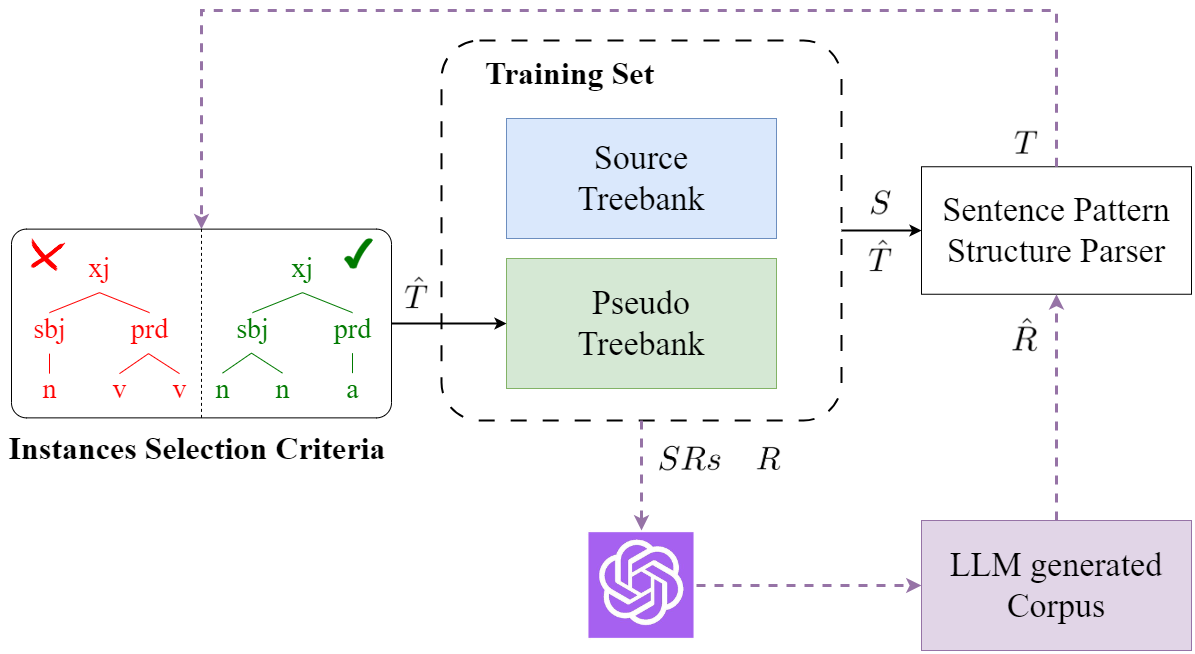}
    \caption{LLM-enhanced self-training frameworks for cross-domain SPS parsing.}
    \label{fig3}
\end{figure}

\subsection{LLM-enhanced Self-training}

To improve data diversity, we supply LLMs with syntactic rules and example sentences extracted from the target domain, facilitating the generation of candidate raw corpora for each iteration of self-training.
As the iterative process progresses, the parser's capability gradually improves. As shown in the Figure~\ref{fig3}, the process specifically includes the following steps:

\begin{itemize}
    \item Extract syntactic rules $SRs$ based on existing training data.
    \item Input $SRs$ and target domain example sentences $R$ as prompts into the LLMs to generate raw, domain-specific corpora.
    \item Train a parser $P$ based on source data $S$ and high-quality pseudo-trees $\hat{T}$, which are absent if it is the first round of training.
    \item Utilize the newly trained $P$ to parse the LLM-generated sentences $\hat{R}$, resulting in the parsed output $T$.
    \item Employ the Instances Selection Criteria to select the top $K$ pseudo-trees to be added to $\hat{T}$.
\end{itemize}

To generate higher-quality sentences that more closely resemble the target domain, we constrained the prompts for LLMs from three aspects: syntactic rules, example sentences, and length. We introduced a Gaussian distribution for the average length of source domain data and randomly sampled values at each request to constrain the length of the generated sentences, as well as the number of required syntactic rules. 

\begin{figure}[t]
    \centering
    \includegraphics[width=1\linewidth]{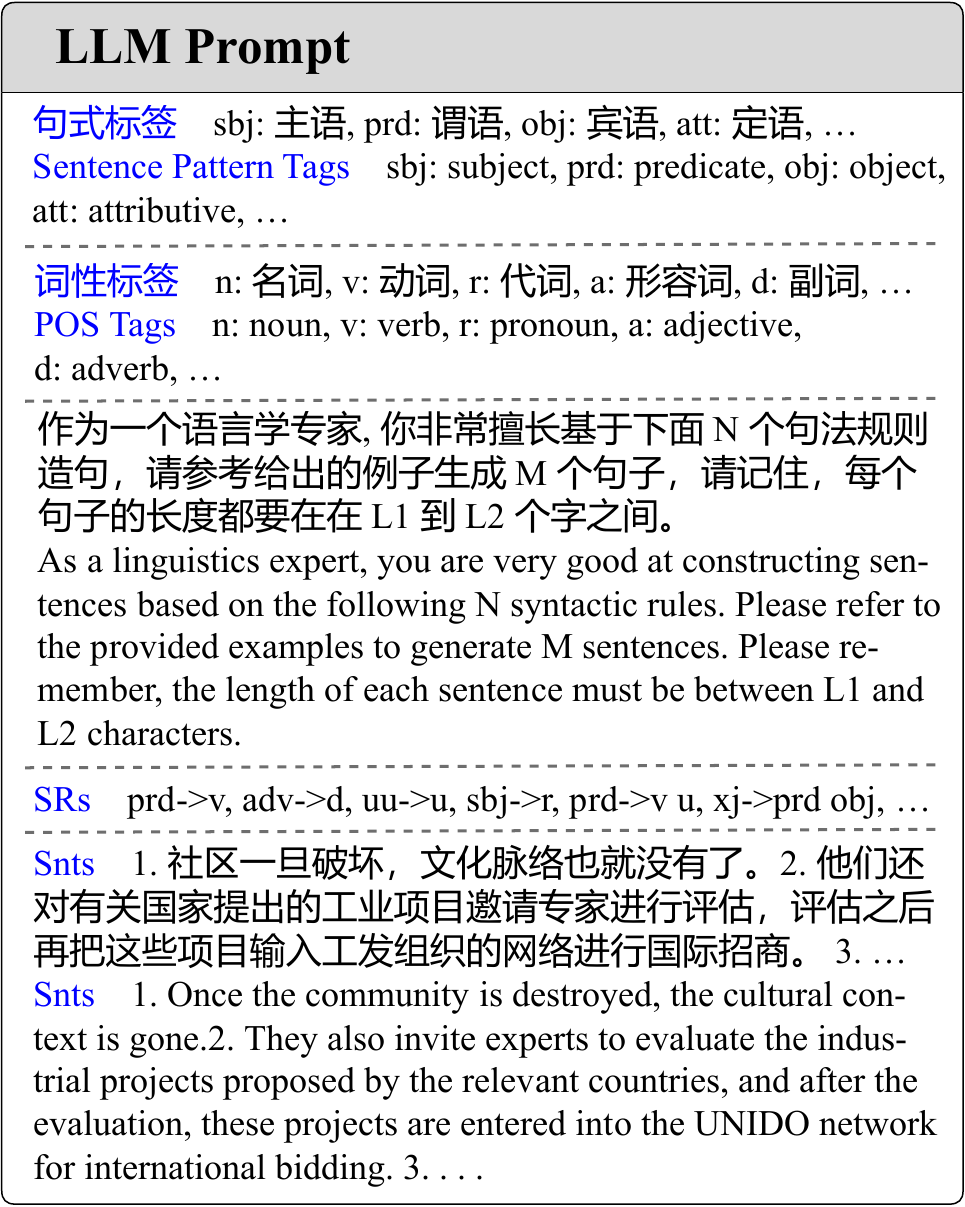}
    \caption{LLMs prompts example for generating sentences based on syntactic rules and target domain instances. Note that the blue markers and dotted lines are not components of the actual prompt but are included solely for illustrative purposes.}
    \label{fig4}
\end{figure}

The example of the LLM prompt is illustrated in Figure~\ref{fig4}.

To verify whether LLMs genuinely adhere to our constraints during data generation, we randomly extracted 20 sentences of data from the raw corpus generated by LLMs. Then manual annotation and matched them with the syntactic rules provided in the prompt. We found that the syntactic structure of the generated sentences matched the provided syntactic rules 69.3\% of the time. Furthermore, when the prompt includes uncommon labels such as $ind$ (independent word), LLMs tend to generate sentences with the $ind$ label.

\subsection{Rule-based Instances Selection Criteria}

\citet{yang:22} and \citet{wang:23} have confirmed the feasibility of grammar-rule-based selection criteria. To filter higher-quality pseudo-data from the raw corpus generated by the LLM, we propose mapping-rule-based instance selection criteria. Unlike previous self-training selection criteria that focus on grammar rules extracted from the source domain \citep{li:23}, this criterion considers the selection based on syntactic rules from the target domain. 

We first get target domain SPS trees converted via mapping rules. Differently from previously calculating the distance between pseudo-data and training data, we calculate the distance between pseudo-data and the converted target domain data to filter pseudo-data suitable for cross-domain parsing self-training.

The Jensen-Shannon (JS) divergence is a symmetric and finite measure of similarity between two probability distributions, often used to quantify the difference in information content or to assess the similarity between datasets. We employ the JS divergence shown in Equation \eqref{eq:1} to assess the similarity between the target data converted by mapping rules and individual instances:

\begin{equation}
    D(c, S) = JS(S, S \cup {c})
\label{eq:1}
\end{equation}
\begin{equation}
    \text{instances} = \text{Top-}K \mathop{\arg\min}_{c \in C} D(c, S)
\label{eq:2}
\end{equation}
where $S$ represents the target domain dataset transformed by rules, and $C$ is the candidate set, as well as the candidate instances are denoted as $c \in C$. 

We then select the top $K$ candidate sentences as shown in Equation \eqref{eq:2}. These sentences possess syntactic structures more closely aligned with the converted target domain data and serve as an extension of the training set in subsequent iterations of self-training to achieve a gradual adaptation to the target domain.

Our criteria has three levels: token, confidence, and syntactic rule. Additionally, we considered the integration of criteria based on confidence and syntactic rules to establish a more effective standard, aimed at better filtering instances that adapt to the target domain, whick is also supported by \citet{li:23}.

\textbf{Token-based criteria} filter sentences whose token distribution is closer to that of the source domain treebank (using the JS distance as in equation \eqref{eq:2}).

\textbf{Conf-based criteria} focus on pseudo-trees provided by the model with high confidence. 

\textbf{SRs-based criteria} filter sentences that are structurally more similar to the source data by calculating the JS distance between instances and the set of grammatical rules in the treebank of the source domain.

\textbf{SRsConf-based criteria} combine scores from both Conf-based and SRs-based criteria, and select high-confidence instances among candidates with high scores on the syntactic rule while considering both structural information and data reliability.

\textbf{CSRs-based criteria} differ from the SRs-based criteria in that it compares instances with the set of syntactic rules of the target domain that have undergone rule transformation. 

\textbf{CSRsConf-based criteria} combine Conf-based scores with CSRs-based scores to obtain more effective training data.

\section{Experiment}

\subsection{Data}

We use the STB as our source domain (Textbook) and the CTB as the target domain (News), and employ GPT-4 to generate 40k sentence points for data augmentation in each iteration.

\textbf{CTB.} Penn Chinese Treebank (CTB), a widely used constituent treebank in linguistic research, with existing rule-based transformation studies targeting CTB5 \cite{xie:23}. Following the data segmentation approach of \citet{liu:17} for CTB5, we constructed the training and development sets in a 50:1 ratio, as illustrated in Table~\ref{dataset-split}. The development and test datasets were manually annotated by our annotators.

\textbf{STB.} Sentence Pattern Structure Treebank (STB), constructed by Beijing Normal University, which primarily comprises sentences from internationally influential Chinese language textbooks, comprising approximately 70k sentences in 1,040 articles. The length of sentences in this corpus varies significantly, with the shortest comprising only 2 tokens and the longest extending to 217 tokens. Additionally, sentences within news corpora tend to be lengthy; for instance, the average length in CTB5 is 362. Therefore, while retaining complete articles, we delete articles containing short sentences, ultimately preserving 10k sentences across 785 articles, with an average sentence length of 32. To enhance the relevance and comparability of our evaluation results, we adopted the data segmentation approach for CTB5 by \citet{liu:17}, as shown in Table~\ref{dataset-split}.

\renewcommand{\arraystretch}{1.2} 
\begin{table}[t]
\centering
\begin{minipage}{0.5\textwidth}
\centering
\begin{tabular}{>{\centering\arraybackslash}m{45pt} >{\centering\arraybackslash}m{40pt} >{\centering\arraybackslash}m{25pt} >{\centering\arraybackslash}m{25pt}}
\specialrule{1pt}{0pt}{0pt}
\textbf{Dataset} & \textbf{train} & \textbf{dev} & \textbf{test}\\
\hline
CTB5 & 17,544 & 352 & 318\\
\hline
STB & 9,700 & 193 & 193\\
\specialrule{1pt}{0pt}{0pt}
\end{tabular}
\caption{STB and CTB Data Statistics}
\label{dataset-split}
\end{minipage}
\end{table}

\begin{figure}[t]
\centering
\begin{tikzpicture}
    \centering
    \Tree [.adv [.\node(root){t}; 
                    [.t [.昨天 \edge[dashed]; last ] ]
                    [.t [.晚上 \edge[dashed]; night ] ] ]
                [.w ， ] ]
    \begin{scope}[shift={(1.5in,0)}]
    \Tree [.adv [.\node(site){t}; [.昨天 \edge[dashed]; last ] ]
                [.t [.晚上 \edge[dashed]; night ] ]
                [.w ， ] ]
    \end{scope}
    \draw[->, red, dashed, line width=2pt] (root) to[bend right=45] (site);
\end{tikzpicture}
    \caption{Example of removing redundant non-leaf node POS tags. Note that in the SPS grammar, punctuation marks are treated as suffixes to the preceding word.}
    \label{fig5}
\end{figure}

Due to the annotation standards of STB not aligning with CTB, we make a unification of the format. Initially, the word segmentation granularity in the STB was finer, including a broader range of colloquial words, idioms, and slang that are not present in the News domain, necessitating a word segmentation granularity transformation. Additionally, not all part-of-speech (POS) tags in the STB are located at leaf nodes as shown in Figure~\ref{fig5}, which significantly affects the accuracy of evaluations. Therefore, we removed all extraneous non-leaf node POS tags.

\begin{table}[t]
\centering
\begin{tabular}{>{\centering\arraybackslash}m{65pt} >{\centering\arraybackslash}m{60pt} >{\centering\arraybackslash}m{60pt}}
\specialrule{1pt}{0pt}{0pt}
 & \textbf{Before} & \textbf{After}\\
\hline
\textbf{Merge} & 圣诞\ \ 节 & 圣诞节\\
\hline
\textbf{Split} & 武侠小说 & 武侠\ \ 小说\\
\hline
\textbf{Misalignment} & 惹火\ \ 儿了 & 惹\ \ 火儿\ \ 了\\
\specialrule{1pt}{0pt}{0pt}
\end{tabular}
\caption{Three scenarios requiring word segmentation granularity transformation.}
\label{three-cases}
\end{table}

\textbf{Word Segmentation Granularity Transfer.} Due to the differing word segmentation granularity between the STB as the source domain and the CTB as the target domain, considering the universality of the CTB corpus, we designed an algorithm to give a word segmentation granularity transformation on the STB. This process aligns its word segmentation granularity with that of the CTB, ensuring uniformity between the two.

As shown in Table~\ref{three-cases}, we deal with three scenarios: merge, split, and misalignment. The merge scenario involves combining words in the STB that are of finer granularity than those in the CTB, primarily utilizing the CTB lexicon to conduct prefix matching with words in the STB to determine if merging is necessary. The split scenario involves dividing words in the STB that are of coarser granularity than those in the CTB, mainly using dynamic words, or out-of-vocabulary words, in the STB. The STB annotates the internal structure of dynamic words. For words that require splitting, we refer to the dynamic words' splitting method. Lastly, for a small number of misalignment cases, where the word segmentation standards for the same phrase differ or dynamic words cannot address the split, we resort to manual annotation for resolution.

The process of word segmentation granularity transformation is iterative. Initially, we split the STB into the finest granularity based on the splitting method for dynamic words. Next, the merging process is as follows: \textbf{1) Prefix Matching:} Upon getting the leaf nodes in the syntax tree, we first determine if this leaf node is a prefix of a word in the CTB lexicon. If it is a prefix and shares the same parent with the next leaf node, a merging match is attempted. If the merging word exists in the lexicon, combine them, if not, this may indicate a misalignment case. \textbf{2) Lexicon Matching:} If the leaf node is not a prefix, we then determine whether it exists directly in the lexicon. If it does, we proceed without action. \textbf{3) Error Logging:} If a leaf node is neither a prefix nor exists in the lexicon, the term may represent a scenario unmanageable by dynamic words and is thus documented.

Due to the fact that many words in the CTB lexicon can be both prefixes and complete words, such as ``中国'' (China), which can stand alone as a word or serve as a prefix in ``中国人" (Chinese people), we filter sentences that did not match in the above process. In this round, we first determine whether the obtained word is a lexicon word and then assess its potential as a prefix. Finally, 766 sentences were identified for annotation.

\begin{table}[t]
\centering
\begin{tabular}{ccccc}
\specialrule{1pt}{0pt}{0pt}
\textbf{Method} & \textbf{Textbook} & \textbf{News}\\
\hline
\citet{kitaev:18} w/o & 91.75 & 67.84\\
\hline
\citet{kitaev:18} w/ & 88.01 & 73.61\\
\specialrule{1pt}{0pt}{0pt}
\end{tabular}
\caption{\label{segmentation-result}
Comparison of the effectiveness of training the SPS parser with data before and after the application of word segmentation granularity transfer.
}
\end{table}

As shown in Table~\ref{segmentation-result}, our transformation of the granularity of the word segmentation significantly aids the performance across the domain, resulting in a 5.77-point increase in the news domain.

\subsection{Parameters}

We adopt the self-training method for our parser, which is consistent with the approach utilized by \citet{kitaev:18}. During the self-training iterations, the raw corpus is tokenized by CoreNLP and subsequently tagged by the trained parser. The self-training procedure encompasses four iterations in all cases, with a selection of the top 2k pseudo-trees from a pool of 10k examples generated by the LLMs during each iteration. These selected instances are then integrated into the training set for subsequent iterations. For the LLM-enhanced SPS parser, we extract grammatical rules from both available treebanks and those generated by rules, integrating them with GPT-4 for the generation of raw corpora. All parsers utilize three distinct seeds and the performance is measured as the average F1 score.

\begin{table*}
\centering
\begin{tabular}{>{\centering\arraybackslash}m{150pt} >{\centering\arraybackslash}m{75pt} >{\centering\arraybackslash}m{80pt} >{\centering\arraybackslash}m{75pt}}
\specialrule{1pt}{0pt}{0pt}
\textbf{Method} & \textbf{Criteria} & \textbf{Textbook} & \textbf{News}\\
\hline
Rule-based Mapping & - & - & 73.36\\
\hline
Rule-based Parser & - & 70.87 & 73.51\\
\hline
\citet{kitaev:18} & - & 88.01 & 73.61\\
\hline
\multirow{6}{150pt}{\arraybackslash\parbox{150pt}{\centering LLM-enhanced Self-training \\ w/ Rule-based Criteria}} & token & 87.94 (-0.07) & 75.22 (+1.61)\\
 & SRs & 88.31 (+0.30) & 75.25 (+1.64)\\
 & Conf & 88.32 (+0.31) & 74.63 (+1.02)\\
 & SRs-Conf & 88.04 (+0.03) & 74.51 (+0.90)\\
 & \textbf{CSRs} & \textbf{88.38 (+0.37)} & \textbf{75.29 (+1.68)}\\
 & CSRs-Conf & 88.29 (+0.28) & 74.91 (+1.30)\\
\hline
\multirow{3}{150pt}{\arraybackslash\parbox{150pt}{\centering LLM-enhanced Self-training \\ w/ Rule-based Mapping}} & token & 88.05 (+0.04) & 73.20 (-0.41)\\
 & SRs & 88.26 (+0.25) & 74.17 (+0.56)\\
 & CSRs & 88.21 (+0.20) & 73.96 (+0.35)\\
\specialrule{1pt}{0pt}{0pt}
\end{tabular}
\caption{\label{main-result}
Main results of Rule-based Mapping, Rule-based SPS Parser and Rule-based LLM-enhanced Self-training (ST) with six pseudo-data selection criteria:Token, Conf, GRs, GRsConf, CRGRs and CRGRsConf.
}
\end{table*}

\begin{figure*}[t]
    \centering
    \includegraphics[width=1\linewidth]{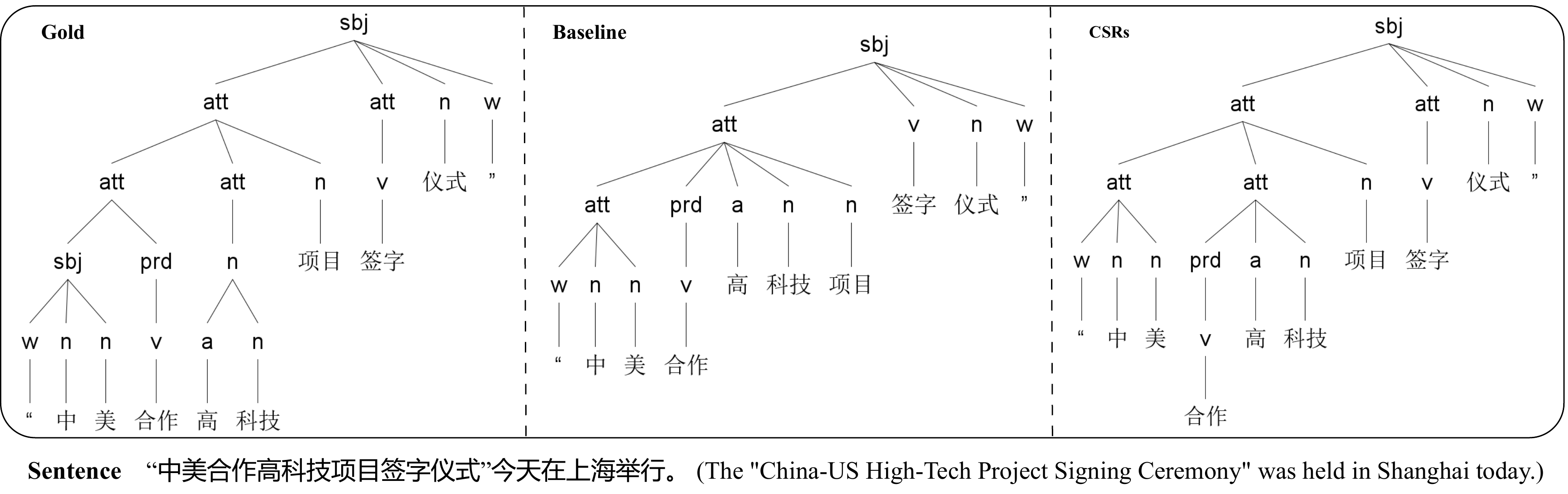}
    \caption{Examples of SPS parsing by the baseline and our method. The left side shows the gold standard, the middle displays the results of baseline parsing, and the right side presents the results parsed by the method of LLM-enhanced Self-training + CRS-based criteria.}
    \label{fig7}
\end{figure*}

\subsection{Main Results}

The principal comparative study was conducted employing the bert-base-chinese model, with experiments carried out on data post-harmonization of word segmentation granularity. The efficacy of the SPS parser across both source and target domains is delineated in the accompanying Table~\ref{main-result}.

In the first stage of our experiment, we implemented a rule-based approach to transform the test set of the target domain. We first utilized a constituency parser to generate constituent trees. Subsequently, leveraging the rule-based mapping relationships between the constituent treebank and the SPS treebank produced SPS trees. Testing within the target domain yielded an F1 score of 73.36. The direct application of rules to generate SPS trees, however, still resulted in issues concerning non-leaf nodes being parts of speech, leading to suboptimal outcomes.

Secondly, we converted all constituent trees from CTB into SPS trees, which were then employed to train a Rule-based SPS Parser. From the result, the method of training the parser offers a marginally better performance in the target domain compared to the direct application of rule-based conversion, with the F1 score increasing from 73.36 to 73.51. However, the rule-based parser exhibits limitations due to the inflexibility of the rule-generated training set, which reduces the diversity of the corpus and leads to a significant decline in performance within the source domain, with an F1 score of only 70.87. Furthermore, in instances where the rules fail to cover, the model may reinforce errors during the training process, resulting in decreased accuracy.

In addition, we explored the efficacy of direct model transfer for cross-domain SPS parsing, positioning it as a robust baseline method compared with LLMs’ parsing. The parser was trained on the processed STB data and subsequently applied directly to the target domain. The results revealed a significant discrepancy between the source and target domains, quantified as a 88.01 - 73.61 = 14.4 point. As mentioned in Section 4.1, there is a significant discrepancy between the data in the source and target domains. The average sentence length in the textbooks is 32, whereas in CTB data, the average sentence length extends to 362. On the other hand, the source domain data from international Chinese textbooks are considerably less challenging than data from the news domain and contain biases, posing a cross-domain challenge.

Lastly, we explored LLM-enhanced self-training for SPS parsers, employing the six selection strategies. Notably, the GRs-based selection shows a bit more enhancement compared to the Conf-based selection. This further illustrates that the effectiveness of the selection criteria is significantly influenced by the quality of the raw corpus utilized in self-training. The positive results also validate the efficacy of our approach, which employs LLMs to generate target domain sentences in each iteration. Compared to the basic model transfer, our LLM-enhanced method achieves an improvement of 1.68. Importantly, it does not compromise the parser's performance in the source domain, on the contrary, it results in an enhancement of 0.37. Moreover, the method of rule-based conversion applied to instance selection achieves better results, indicating that the SPS trees generated by rules possess higher structural accuracy and are closer to the target domain. Employing rule-based methods in instance filtering also better leverages the structural advantages of data, while simultaneously minimizing the adverse effects caused by a lack of flexibility and potential errors within the data.

\subsection{Comparative Results}

As shown in Table~\ref{main-result}, we also conducted another set of experiments on LLM-enhanced self-training. Unlike the experiments solely based on rule-based criteria, in this set, we additionally utilized rule-based mapping. In previous approaches, we employed an existing parser to analyze the raw corpora generated by LLMs and selected the $top K$ pseudo trees. In the approach utilized rule-based mapping, we apply a rule-mapping method to parse the raw corpora. Specifically, this method involves initially conducting constituency parsing on the raw corpora, followed by transforming it into SPS trees through a rule-mapping algorithm, and subsequently choosing the $top K$ pseudo trees. Since this method does not involve the use of models for parsing, it is not possible to utilize confidence-related criteria.

Overall, the performance of LLM-enhanced Self-training combined with Rule-based Mapping significantly lags behind methods that employ parsing. This discrepancy is evident when considering the Token-based criterion. If the Token-based method is applied, the examples selected by the two experiments of LLM-enhanced Self-training are identical, yet the outcomes differ by 2.02 points. This difference clearly demonstrates that the parser trained by our method achieves a much higher accuracy compared to rule-based mapping. Furthermore, during the iterative process, tree conversion by rule is prone to cumulative errors. This is observable from the results of using SRs-based and CSRs-based selection methods, indicating that increased reliance on rules during the training process tends to deteriorate the result.

In summary, whether it involves reliance on LLMs or on rules, the training process should adhere to the principle of moderation. We have leveraged the generative capabilities of LLMs while avoiding the issues of insufficient flexibility and error accumulation caused by rule dependency, which can get more improvement.

\subsection{Case Study}

We use CSRs-based criteria as an example to demonstrate how the parser developed through our method exhibits enhanced domain adaptability. As evidenced in Section 4.1, the data from the source domain, which pertains to Chinese as a Foreign Language education, feature simpler vocabulary, shorter sentence lengths, and more unitary sentence structures. Conversely, the target domain data, derived from news, contains longer sentences with more complex structures. Therefore, a parser trained on news domain data is better equipped to handle complex sentences. In contrast, the baseline trained solely on textbook corpora lack the capability to manage complex sentence structures effectively. As shown in Figure~\ref{fig7}, our model demonstrates superior capability in handling "att", which stands for attributive labels. For example, our model is adept at identifying ``高科技 (high-tech)'' as the attributive of ``项目 (project)'' and ``签字 (signing)'' as the attributive of ``仪式 (ceremony)'', whereas the baseline model cannot achieve this.

\section{Conclusion}

In this study, we develop a novel approach that combines rule-based criteria with the generative capabilities of LLMs for cross-domain adaptation in SPS parsing. By harnessing the generation of LLMs and integrating them into the self-training process, we showed that our approach considerably enhances the performance of the cross-domain SPS parsing. Our method effectively leverages the rule-based selection criteria, gradually moving the training data closer to the target domain. Through experiments, we have validated the efficacy of our methodology on cross-domain tasks and improved performance in the target domain. In conclusion, our rule-based LLM-enhanced self-training approach offers promising solutions for cross-domain adaptation tasks.

\section*{Limitations}

Due to the widespread application of news data, this work initially selects the news domain as the target domain, with plans to extend our exploration to additional domains in subsequent studies. Furthermore, future efforts will involve comparing existing real corpora with the LLM-enhanced method. There are many detailed explorations for LLM-equipped self-training in the raw corpus generation partition, e.g. the influence of different prompts. For time constraints, was limited to the final results. Future analysis should also consider the influence of various factors during each iteration.


\bibliography{acl_latex}



\end{CJK*}
\end{document}